\documentclass{article} 
\usepackage{iclr2023_conference_tinypaper,times}
\usepackage{graphicx}

\usepackage{amsmath,amsfonts,bm}

\def\eqref#1{equation~\ref{#1}}

\def\1{\bm{1}}

\DeclareMathAlphabet{\mathsfit}{\encodingdefault}{\sfdefault}{m}{sl}
\SetMathAlphabet{\mathsfit}{bold}{\encodingdefault}{\sfdefault}{bx}{n}

\usepackage{tabularx}
\usepackage{multirow}
\usepackage{hyperref}
\usepackage{url}
\usepackage{amsmath}
\usepackage{booktabs}
\usepackage{xcolor}

\title{Multilingual Prosody Transfer: Comparing Supervised \& Transfer Learning}

\author{Arnav Goel\thanks{Equal Contribution},\hspace{2mm}Medha Hira\footnotemark[1] \hspace{0.25mm} \& Anubha Gupta \\
Indraprastha Institute of Information Technology\\
New Delhi - 110020, India \\
\texttt{\{arnav21519,medha21265,anubha\}@iiitd.ac.in} \\
}

\iclrfinalcopy 
\begin{document}

\maketitle

\begin{abstract}
The field of prosody transfer in speech synthesis systems is rapidly advancing. This research is focused on evaluating learning methods for adapting pre-trained monolingual text-to-speech (TTS) models to multilingual conditions, i.e., Supervised Fine-Tuning (SFT) and Transfer Learning (TL). This comparison utilizes three distinct metrics: Mean Opinion Score (MOS), Recognition Accuracy (RA), and Mel Cepstral Distortion (MCD). Results demonstrate that, in comparison to SFT, TL leads to significantly enhanced performance, with an average MOS higher by 1.53 points, a 37.5\% increase in RA, and approximately, a 7.8-point improvement in MCD. These findings are instrumental in helping build TTS models for low-resource languages.

\end{abstract}
\vspace{-1em}
\section{Introduction}
\vspace{-1em}
Recent advancements in deep learning, as seen in systems such as fairseq-S2T \citep{wang2022fairseq}, SpeechT5 \citep{ao2022speecht5}, and VITS \citep{kim2021conditional} have significantly enhanced speech synthesis, paving the way for our research on controllable Text-to-Speech (TTS) systems that transfer both text and prosody to target audio. Our study focuses on multilingual prosody transfer in TTS, particularly exploring models initially trained in English and then adapted to other languages. Adapting TTS for multilingual use involves various representation learning methods, including semi-supervised and self-supervised learning \citep{10095702}. We assess two key approaches—supervised fine-tuning (SFT) on text-audio pairs and transfer learning (TL)—to evaluate their effectiveness in generating high-quality audio and transferring prosody in multilingual contexts.
\vspace{-1.5em}
\section{Related Work}
\vspace{-1em}
Prosody transfer and voice conversion in TTS systems have evolved from traditional HMMs, RNNs, and CNNs to Transformer-based architectures like VTN \citep{vaswani2023attention, huang2019voice}. Recent techniques include using 1) ASR for linguistic representation \citep{tian2019vocoderfree, popov2022diffusionbased}, 2) speaker-dependent prosody capture \citep{zhang2020voice}, 3) global cues like pitch and loudness \citep{gururani2020prosody}, and 4) combining local and global prosodic features \citep{huang2023holistic}. {While \cite{saeki2023virtuoso} explore semi-supervised learning for TTS in multilingual settings and \cite{shah2023mparrottts} introduce a pre-trained TTS model for low-resource languages}, the use of speaker embeddings for prosody transfer and adapting pre-trained English TTS systems to multilingual contexts remains less explored. Our study aims to fill this gap by investigating these methods.
\vspace{-2.25em}
\section{Dataset}
\vspace{-1.2em}
To conduct experiments in German, French, Spanish, and Dutch, we utilized segments from the VoxPopuli dataset \citep{wang2021voxpopuli}, comprising 282, 211, 166, and 53 hours of transcribed audio, respectively. For Hindi and Tamil, we selected subsets from the Common Voice corpus \citep{ardila2020common}, amounting to 20 and 200 hours of audio of each language. {To address class imbalance, we uniformly sampled data from each language to equalize dataset duration to approximately \textbf{20 hours}}.

\section{Methodology}
\vspace{-1em}
We aim to adapt pre-trained models for multilingual prosody preservation using Supervised Fine-Tuning (SFT) and Transfer Learning (TL). Our experimental setups and methodologies are as below:

\textbf{SFT:} We selected SpeechT5 for SFT due to its encoder-decoder structure that generates mel-spectrograms from text input. Its audio post-net can incorporate speaker embeddings for prosody transfer \citep{ao2022speecht5}. We utilized x-vector embedding \citep{8461375}, known for capturing emotional and gender characteristics in speech embedding. {The choice of x-vectors is based on experiments detailed in \ref{a_se}}. These embedding are integrated with the output of SpeechT5's decoder to preserve prosody. SpeechT5 is fine-tuned using supervised learning on (text, spectrogram) data pairs aiming to minimise cross-entropy loss. {The implementation details and plots are shown in \ref{fine_tune}}. {This technique adapts a pre-trained monolingual model to multilingual settings.}

\textbf{TL:} To evaluate TL, we implemented the voice cloning method from \cite{jia2019transfer}. This involved a pre-trained speaker encoder designed for speaker identification, merged with a voice conversion model. {For speech synthesis from text, we used MMS TTS \citep{pratap2023scaling} which is a pre-trained multilingual model which does not preserve prosody}. Concurrently, x-vector embedding were derived using a pre-trained encoder \citep{ravanelli2021speechbrain} from input audio. The synthesized audio from MMS and the embedding from the encoder were fed into FreeVC's pre-trained voice conversion module \citep{li2022freevc} to produce prosody-preserving audio.

Our experiment involved regenerating input audio clips using the two described models. We assessed the quality of these generated clips using MOS, RA, and MCD. MOS evaluates the naturalness, quality, and prosody transfer of the generated speech, while RA assesses the respondents' ability to recognize the speaker in both the original and generated audio. MOS and RA are computed based on feedback from 35 respondents\footnote{Detailed calculation methodology and protocol are provided in Appendix \ref{a}}. Although MCD is not an ideal metric for speech quality, it is useful in this context for measuring the distortion between the generated audio clip and the original, thereby aiding in evaluating our experiment's outcomes.

\renewcommand{\arraystretch}{1}
\begin{table}[t]
\label{bert}
\centering
\caption{Comparative performance of SFT and TL on the synthesised speech quality}
\small 
\begin{tabularx}{\textwidth}{|X|X|X|X|X|X|X|}
\hline
\multirow{2}{*}{\textbf{Language}} & \multicolumn{3}{c|}{\textbf{Supervised Fine-Tuning (SpeechT5)}} & \multicolumn{3}{c|}{\textbf{Transfer Learning (FreeVC)}} \\ \cline{2-7}
 &MOS ($\uparrow$)$^\star$&Recognition Accuracy($\uparrow$)&MCD ($\downarrow$)&MOS ($\uparrow$)$^\star$ &Recognition Accuracy($\uparrow$)&MCD ($\downarrow$)\\ 
\hline
Spanish & 2.73 $\pm$ 0.03 & 0.43 & 23.23 & 4.11 $\pm$ 0.12 & 0.81 & 15.83 \\
\hline
French & 2.85 $\pm$ 0.09 & 0.48 & 21.36 & 4.26 $\pm$ 0.09 & 0.83 & \textbf{12.54} \\
\hline
German & 3.01 $\pm$ 0.13 & 0.52 & 20.08 & \textbf{4.35 $\pm$ 0.01} & \textbf{0.88} & 13.41 \\
\hline
Dutch & 2.44 $\pm$ 0.05 & 0.45 & 24.74 & 4.15 $\pm$ 0.04 & 0.79 & 17.21 \\
\hline
Hindi & 2.32 $\pm$ 0.06 & 0.40 & 25.23 & 4.01 $\pm$ 0.17 & 0.82 & 16.28 \\
\hline
Tamil & 2.12 $\pm$ 0.24 & 0.37 & 26.21 & 3.85 $\pm$ 0.13 & 0.77 & 18.54 \\
\hline
\end{tabularx}
\flushleft
\small{$^\star$mean$\pm$std} 
\vspace{-2em}
\end{table}
\label{tab:results}
\renewcommand{\arraystretch}{1}
\vspace{-1em}
\section{Results}
\vspace{-1em}

Table 1 shows the results of our experiments. Audio clips generated by SpeechT5-finetuned using SFT are generally found to be noisy and have poor audio quality. This is further validated by the MOS scores which are reported at a \emph{95\% confidence level}. TL surpasses SFT by an average of 1.53 points over the six languages. Additionally, the recognition accuracy of the TL generated audio exceeds that of SFT by more than 35\% on average. These scores substantiate that Transfer Learning is superior in retaining the unique characteristics of a voice. While adapting the model to another language, SFT reduces the model's ability to generate good quality speech, let alone preserve prosody. MCD compares the mel-frequency cepstral coefficients (MFCC) of ground truth and generated speech. We used dynamic time-warping to calculate MCD in order to account for clips with varying lengths. TL yields lower MCD compared to SFT (indicating closer resemblance). The distortion is lesser by an average of 35\% on all the six languages. 
\vspace{-1.25em}
\section{Conclusion and Future Work}
\vspace{-1em}
Our findings highlight the superiority of transfer learning over supervised fine-tuning in adapting pre-trained models for TTS applications. This insight is particularly crucial for developing TTS models in low-resource environments, where supervised fine-tuning's data-intensive nature can be a significant challenge. Future research will aim to establish a framework for comparing different learning methods in adapting pre-trained models to low-resource and multilingual contexts.
\vspace{-1.25em}
\subsubsection*{URM Statement}
The authors acknowledge that at all the authors of this work meet the URM criteria of ICLR 2024 Tiny Papers Track.

\bibliography{sbi_paper2}

\begin{thebibliography}{21}
\providecommand{\natexlab}[1]{#1}
\providecommand{\url}[1]{\texttt{#1}}
\expandafter\ifx\csname urlstyle\endcsname\relax
  \providecommand{\doi}[1]{doi: #1}\else
  \providecommand{\doi}{doi: \begingroup \urlstyle{rm}\Url}\fi

\bibitem[Ao et~al.(2022)Ao, Wang, Zhou, Wang, Ren, Wu, Liu, Ko, Li, Zhang, Wei, Qian, Li, and Wei]{ao2022speecht5}
Junyi Ao, Rui Wang, Long Zhou, Chengyi Wang, Shuo Ren, Yu~Wu, Shujie Liu, Tom Ko, Qing Li, Yu~Zhang, Zhihua Wei, Yao Qian, Jinyu Li, and Furu Wei.
\newblock Speecht5: Unified-modal encoder-decoder pre-training for spoken language processing, 2022.

\bibitem[Ardila et~al.(2020)Ardila, Branson, Davis, Henretty, Kohler, Meyer, Morais, Saunders, Tyers, and Weber]{ardila2020common}
Rosana Ardila, Megan Branson, Kelly Davis, Michael Henretty, Michael Kohler, Josh Meyer, Reuben Morais, Lindsay Saunders, Francis~M. Tyers, and Gregor Weber.
\newblock Common voice: A massively-multilingual speech corpus, 2020.

\bibitem[Cao et~al.(2014)Cao, Cooper, Keutmann, Gur, Nenkova, and Verma]{cao2014crema}
Houwei Cao, David~G Cooper, Michael~K Keutmann, Ruben~C Gur, Ani Nenkova, and Ragini Verma.
\newblock {CREMA-D}: Crowd-sourced emotional multimodal actors dataset.
\newblock \emph{IEEE transactions on affective computing}, 5\penalty0 (4):\penalty0 377--390, 2014.

\bibitem[Gururani et~al.(2020)Gururani, Gupta, Shah, Shakeri, and Pinto]{gururani2020prosody}
Siddharth Gururani, Kilol Gupta, Dhaval Shah, Zahra Shakeri, and Jervis Pinto.
\newblock Prosody transfer in neural text to speech using global pitch and loudness features, 2020.

\bibitem[Huang et~al.(2019)Huang, Hayashi, Wu, Kameoka, and Toda]{huang2019voice}
Wen-Chin Huang, Tomoki Hayashi, Yi-Chiao Wu, Hirokazu Kameoka, and Tomoki Toda.
\newblock Voice transformer network: Sequence-to-sequence voice conversion using transformer with text-to-speech pretraining, 2019.

\bibitem[Huang et~al.(2023)Huang, Peloquin, Kao, Wang, Gong, Salesky, Adi, Lee, and Chen]{huang2023holistic}
Wen-Chin Huang, Benjamin Peloquin, Justine Kao, Changhan Wang, Hongyu Gong, Elizabeth Salesky, Yossi Adi, Ann Lee, and Peng-Jen Chen.
\newblock A holistic cascade system, benchmark, and human evaluation protocol for expressive speech-to-speech translation, 2023.

\bibitem[Jia et~al.(2019)Jia, Zhang, Weiss, Wang, Shen, Ren, Chen, Nguyen, Pang, Moreno, and Wu]{jia2019transfer}
Ye~Jia, Yu~Zhang, Ron~J. Weiss, Quan Wang, Jonathan Shen, Fei Ren, Zhifeng Chen, Patrick Nguyen, Ruoming Pang, Ignacio~Lopez Moreno, and Yonghui Wu.
\newblock Transfer learning from speaker verification to multispeaker text-to-speech synthesis, 2019.

\bibitem[Kim et~al.(2021)Kim, Kong, and Son]{kim2021conditional}
Jaehyeon Kim, Jungil Kong, and Juhee Son.
\newblock Conditional variational autoencoder with adversarial learning for end-to-end text-to-speech, 2021.

\bibitem[li et~al.(2022)li, tu, and xiao]{li2022freevc}
Jingyi li, Weiping tu, and Li~xiao.
\newblock Freevc: Towards high-quality text-free one-shot voice conversion, 2022.

\bibitem[Popov et~al.(2022)Popov, Vovk, Gogoryan, Sadekova, Kudinov, and Wei]{popov2022diffusionbased}
Vadim Popov, Ivan Vovk, Vladimir Gogoryan, Tasnima Sadekova, Mikhail Kudinov, and Jiansheng Wei.
\newblock Diffusion-based voice conversion with fast maximum likelihood sampling scheme, 2022.

\bibitem[Pratap et~al.(2023)Pratap, Tjandra, Shi, Tomasello, Babu, Kundu, Elkahky, Ni, Vyas, Fazel-Zarandi, Baevski, Adi, Zhang, Hsu, Conneau, and Auli]{pratap2023scaling}
Vineel Pratap, Andros Tjandra, Bowen Shi, Paden Tomasello, Arun Babu, Sayani Kundu, Ali Elkahky, Zhaoheng Ni, Apoorv Vyas, Maryam Fazel-Zarandi, Alexei Baevski, Yossi Adi, Xiaohui Zhang, Wei-Ning Hsu, Alexis Conneau, and Michael Auli.
\newblock Scaling speech technology to 1,000+ languages, 2023.

\bibitem[Ravanelli et~al.(2021)Ravanelli, Parcollet, Plantinga, Rouhe, Cornell, Lugosch, Subakan, Dawalatabad, Heba, Zhong, Chou, Yeh, Fu, Liao, Rastorgueva, Grondin, Aris, Na, Gao, Mori, and Bengio]{ravanelli2021speechbrain}
Mirco Ravanelli, Titouan Parcollet, Peter Plantinga, Aku Rouhe, Samuele Cornell, Loren Lugosch, Cem Subakan, Nauman Dawalatabad, Abdelwahab Heba, Jianyuan Zhong, Ju-Chieh Chou, Sung-Lin Yeh, Szu-Wei Fu, Chien-Feng Liao, Elena Rastorgueva, François Grondin, William Aris, Hwidong Na, Yan Gao, Renato~De Mori, and Yoshua Bengio.
\newblock Speechbrain: A general-purpose speech toolkit, 2021.

\bibitem[Saeki et~al.(2023{\natexlab{a}})Saeki, Zen, Chen, Morioka, Wang, Zhang, Bapna, Rosenberg, and Ramabhadran]{10095702}
Takaaki Saeki, Heiga Zen, Zhehuai Chen, Nobuyuki Morioka, Gary Wang, Yu~Zhang, Ankur Bapna, Andrew Rosenberg, and Bhuvana Ramabhadran.
\newblock Virtuoso: Massive multilingual speech-text joint semi-supervised learning for text-to-speech.
\newblock In \emph{ICASSP 2023 - 2023 IEEE International Conference on Acoustics, Speech and Signal Processing (ICASSP)}, pp.\  1--5, 2023{\natexlab{a}}.
\newblock \doi{10.1109/ICASSP49357.2023.10095702}.

\bibitem[Saeki et~al.(2023{\natexlab{b}})Saeki, Zen, Chen, Morioka, Wang, Zhang, Bapna, Rosenberg, and Ramabhadran]{saeki2023virtuoso}
Takaaki Saeki, Heiga Zen, Zhehuai Chen, Nobuyuki Morioka, Gary Wang, Yu~Zhang, Ankur Bapna, Andrew Rosenberg, and Bhuvana Ramabhadran.
\newblock Virtuoso: Massive multilingual speech-text joint semi-supervised learning for text-to-speech, 2023{\natexlab{b}}.

\bibitem[Shah et~al.(2023)Shah, Tambrahalli, Kosgi, Pedanekar, and Gandhi]{shah2023mparrottts}
Neil Shah, Vishal Tambrahalli, Saiteja Kosgi, Niranjan Pedanekar, and Vineet Gandhi.
\newblock Mparrottts: Multilingual multi-speaker text to speech synthesis in low resource setting, 2023.

\bibitem[Snyder et~al.(2018)Snyder, Garcia-Romero, Sell, Povey, and Khudanpur]{8461375}
David Snyder, Daniel Garcia-Romero, Gregory Sell, Daniel Povey, and Sanjeev Khudanpur.
\newblock X-vectors: Robust dnn embeddings for speaker recognition.
\newblock In \emph{2018 IEEE International Conference on Acoustics, Speech and Signal Processing (ICASSP)}, pp.\  5329--5333, 2018.
\newblock \doi{10.1109/ICASSP.2018.8461375}.

\bibitem[Tian et~al.(2019)Tian, Chng, and Li]{tian2019vocoderfree}
Xiaohai Tian, Eng~Siong Chng, and Haizhou Li.
\newblock A vocoder-free wavenet voice conversion with non-parallel data, 2019.

\bibitem[Vaswani et~al.(2023)Vaswani, Shazeer, Parmar, Uszkoreit, Jones, Gomez, Kaiser, and Polosukhin]{vaswani2023attention}
Ashish Vaswani, Noam Shazeer, Niki Parmar, Jakob Uszkoreit, Llion Jones, Aidan~N. Gomez, Lukasz Kaiser, and Illia Polosukhin.
\newblock Attention is all you need, 2023.

\bibitem[Wang et~al.(2021)Wang, Rivière, Lee, Wu, Talnikar, Haziza, Williamson, Pino, and Dupoux]{wang2021voxpopuli}
Changhan Wang, Morgane Rivière, Ann Lee, Anne Wu, Chaitanya Talnikar, Daniel Haziza, Mary Williamson, Juan Pino, and Emmanuel Dupoux.
\newblock Voxpopuli: A large-scale multilingual speech corpus for representation learning, semi-supervised learning and interpretation, 2021.

\bibitem[Wang et~al.(2022)Wang, Tang, Ma, Wu, Popuri, Okhonko, and Pino]{wang2022fairseq}
Changhan Wang, Yun Tang, Xutai Ma, Anne Wu, Sravya Popuri, Dmytro Okhonko, and Juan Pino.
\newblock fairseq s2t: Fast speech-to-text modeling with fairseq, 2022.

\bibitem[Zhang et~al.(2020)Zhang, Liu, Chen, Hu, Jiang, Ling, and Dai]{zhang2020voice}
Jing-Xuan Zhang, Li-Juan Liu, Yan-Nian Chen, Ya-Jun Hu, Yuan Jiang, Zhen-Hua Ling, and Li-Rong Dai.
\newblock Voice conversion by cascading automatic speech recognition and text-to-speech synthesis with prosody transfer, 2020.

\end{thebibliography}
\bibliographystyle{iclr2023_conference_tinypaper}

\appendix
\section{Appendix} \label{a}

\subsection{Acronyms Used}

\begin{itemize}
    \item TL : Transfer Learning
    \item SFT : Supervised-Fine Tuning
    \item MCD : Mel Cepstral Distortion
    \item MFCC : Mel-Frequency Cepstral Coefficients
    \item MOS : Mean Opinion Score
    \item ASR : Automated Speech Recognition
    \item TTS : Text to Speech
    \item HMM : Hidden Markov Models
    \item RNN : Recurrent Neural Networks
    \item CNN : Convolutional Neural Networks
    \item MMS : Massively Multilingual Speech \citep{pratap2023scaling}
\end{itemize}
\subsection{Speaker Embeddings} \label{a_se}

X-vector embeddings \citep{8461375}, derived from deep neural networks, excel in capturing intricate speaker characteristics such as emotion and gender. This makes them ideal for prosody transfer tasks, allowing for the addition or alteration of emotional and gender nuances in synthesized or altered speech, thereby improving its naturalness and expressiveness. Our study examines x-vector's efficacy in emotion and gender recognition using the CREMA-D dataset \citep{cao2014crema}. In gender identification, the pink and blue graphs (indicating female and male speakers, respectively) show clear distinction.

After incorporating the x-vector embeddings, we utilized them to train a multi-layer perceptron classifier. This classifier, has two hidden layers of size 128 and 64 respectively, and was trained on 80\% of the data using the cross entropy loss. The ReLU activation function was used for each layer. Testing was conducted for the gender and emotion classification task where X-vectors displayed the highest accuracy.

\begin{table}[h]
\centering
\begin{tabular}{|c|c|c|}
\hline
\textbf{Pre-Trained Embedding} & \textbf{Accuracy (Emotion)} & \textbf{Accuracy (Gender)} \\ \hline
MFCC (13)& 38.04\%& 78.96\%\\ \hline
Wav2Vec2& 46.87\%& 76.29\%\\ \hline
X-vector& \textbf{61.72}\%&\textbf{99.19}\%\\\hline
ECAPA-TDNN& 47.74\%&97.65\%\\\hline
WavLM& 53.27\%&71.32\%\\\hline
\end{tabular}
\caption{Results of pre-trained embeddings on emotion and gender recognition}
\label{table:speaker_embedding_accuracy}
\end{table}

\begin{figure}[h]
    \centering
    \begin{minipage}{0.4\textwidth}
        \centering
        \includegraphics[width=\linewidth]{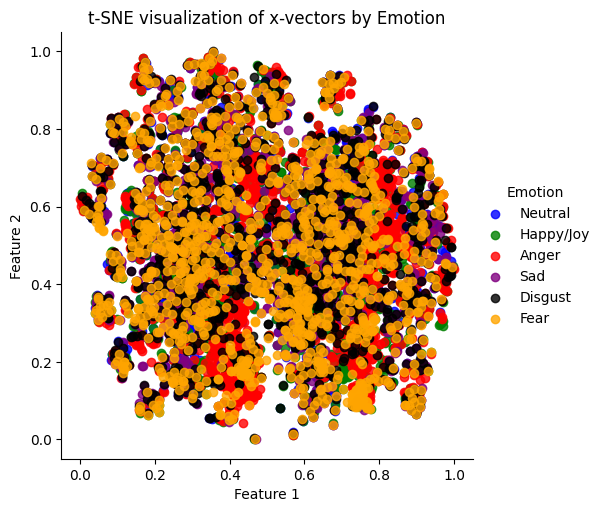}
        \caption{Emotion}
        \label{fig:xvec_emotion}
    \end{minipage}\hfill
    \begin{minipage}{0.4\textwidth}
        \centering
        \includegraphics[width=\linewidth]{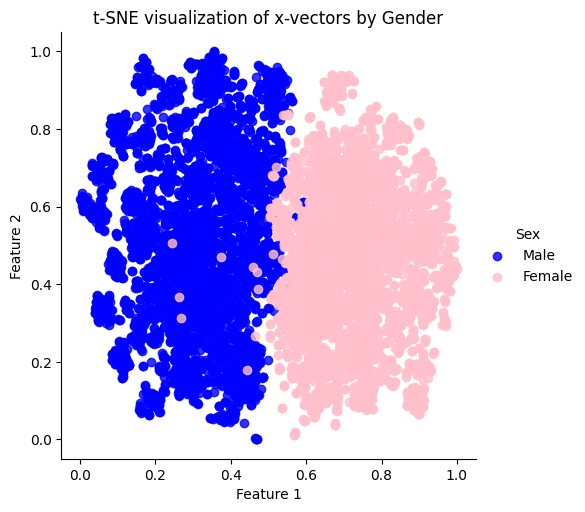}
        \caption{Gender}
        \label{fig:xvec_gender}
    \end{minipage}
    \caption{t-SNE visualisation of x-vector embeddings}
\end{figure}

\subsection{Fine-Tuning Implementation Details and Plots}\label{fine_tune}

\begin{figure}[]
    \centering
    \begin{minipage}{0.68\linewidth}
        \includegraphics[width=\linewidth]{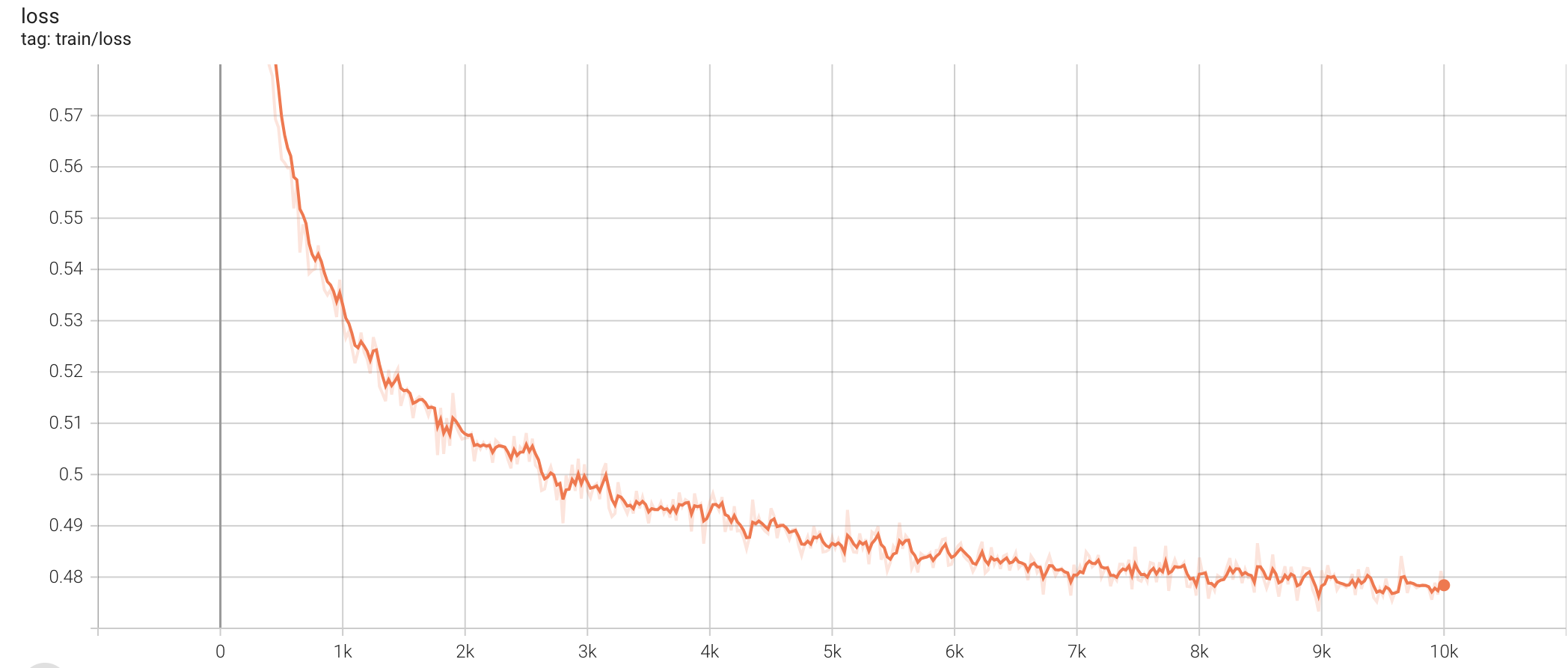}
        \caption{Training Loss vs Epochs for French}
        \label{fig:french_train}
    \end{minipage}
    \hfill 
    \begin{minipage}{0.68\linewidth}
        \includegraphics[width=\linewidth]{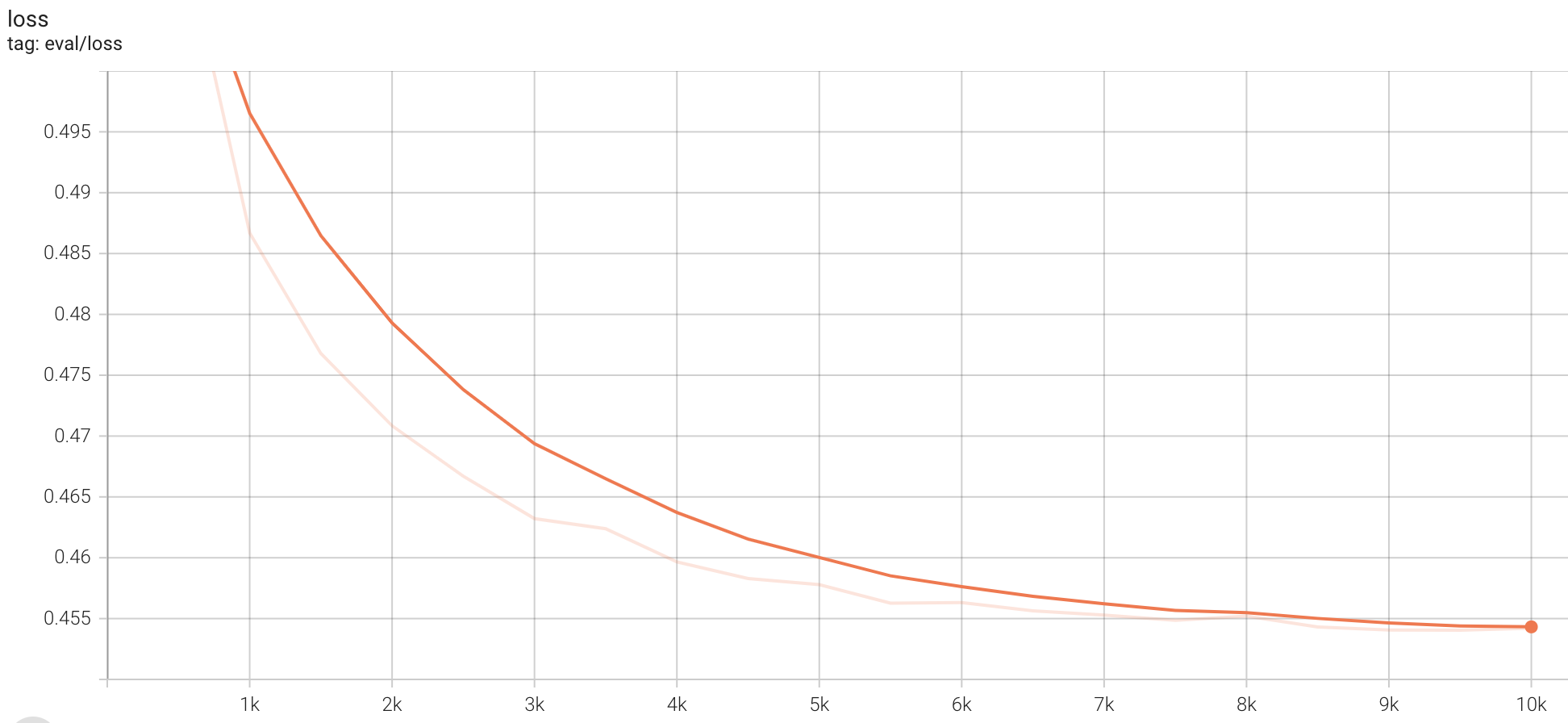}
        \caption{Validation Loss vs Epochs for French}
        \label{fig:french_val}
    \end{minipage}
\end{figure}

The preprocessing details for the data is given as follows:
\begin{enumerate}
    \item Firstly, the dataset is downloaded using an API and the audio files are extracted. Standard audio pre-processing is applied to remove noise and silence before generating speaker embeddings (x-vectors)
    \item Since speakers are annotated for the dataset, the audio clips belonging to speakers with clips in the range of $\in (100, 400)$ are only selected for fine-tuning.
    \item For generating the text transcripts, transliteration is performed by phenome transformation on symbols not existing in English. This is particularly important in Hindi and Tamil where the lexical scripts are entirely different from English and need to be transliterated for fine-tuning to happen.
\end{enumerate}

This preprocessing prepares our dataset for fine-tuning on any English pre-trained TTS after which we train the SpeechT5 model on text-spectrogram pairs. The following hyperparameters are set for fine-tuning SpeechT5: 
\begin{itemize}
    \item \textbf{Learning rate:} $1e-5$
    \item \textbf{Epochs:} 10000
    \item \textbf{Warmup steps:} 500
    \item \textbf{Train Batch Size:} 4
    \item \textbf{Val Batch Size:} 4
    \item \textbf{Gradient accumulation steps:} 8
    \item \textbf{fp16:} True
    \item \textbf{Evaluation Strategy:} "steps"    
\end{itemize}

\begin{figure}[b]
    \centering
    \begin{minipage}{0.48\linewidth}
        \includegraphics[width=\linewidth]{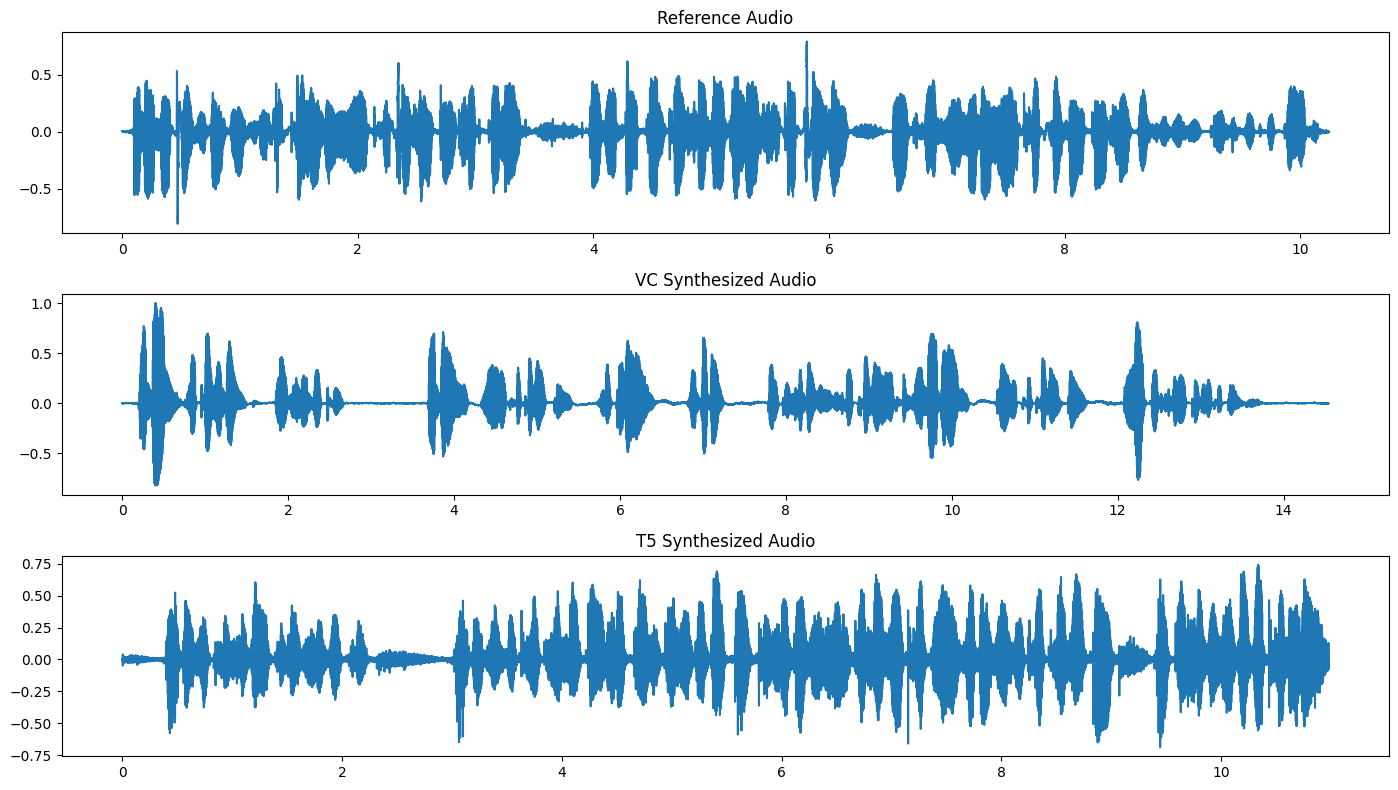}
        \caption{Comparing Waveforms of the three audio clips}
        \label{fig:wav}
    \end{minipage}
    \hfill 
    \begin{minipage}{0.48\linewidth}
        \includegraphics[width=\linewidth]{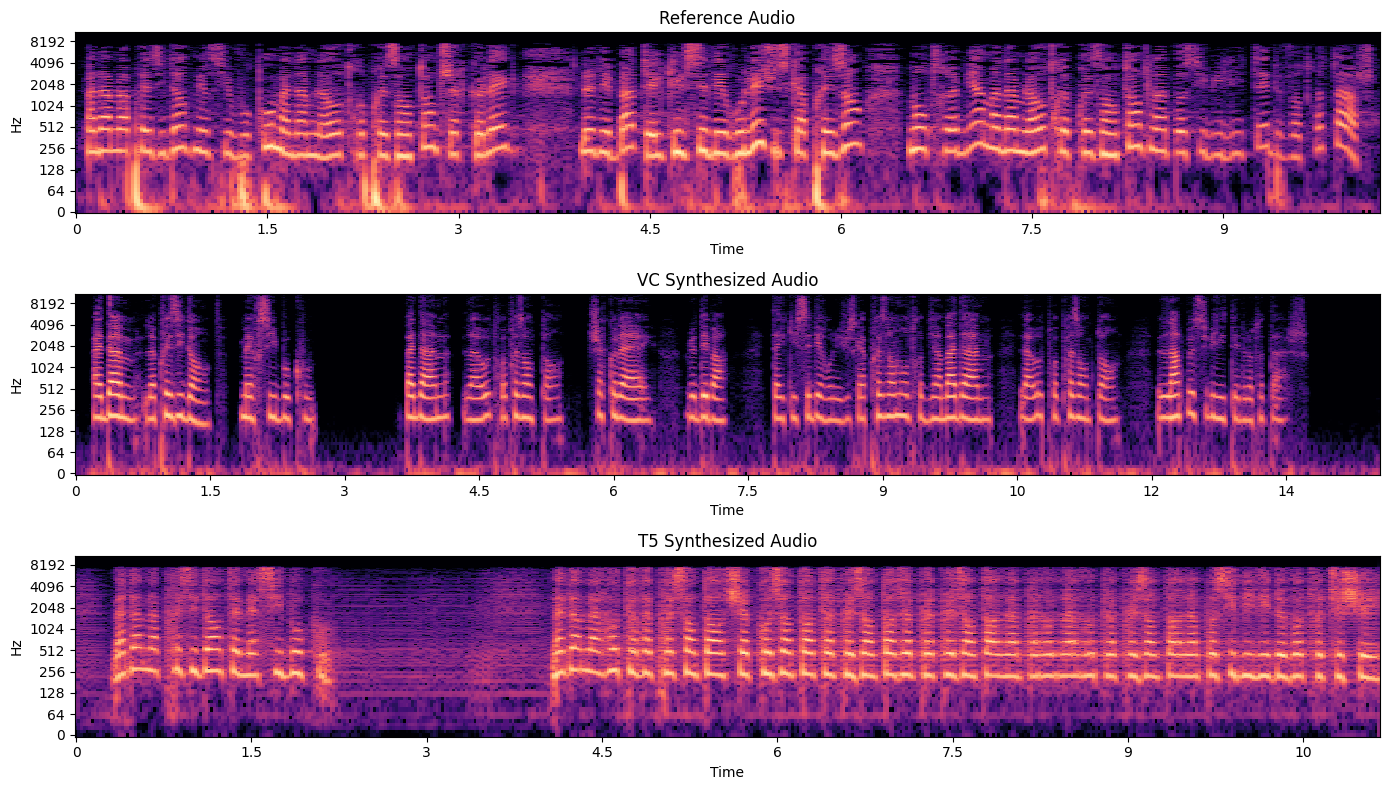}
        \caption{Comparing Spectrograms of the three audio clips}
        \label{fig:spec}
    \end{minipage}
\end{figure}

Figures \ref{fig:french_train} and \ref{fig:french_val} display the loss curves for training and validating SpeechT5 on VoxPopuli-French, converging around 10000 epochs, showing effective model fine-tuning. Figures \ref{fig:wav} and \ref{fig:spec} present a waveform and spectrogram generated by FreeVC and SpeechT5, respectively, using a random test set audio clip, termed \emph{Reference Audio}. This clip, processed for x-vector embeddings, is fed into the fine-tuned SpeechT5, resulting in \emph{T5 Synthesised Audio}, while FreeVC produces \emph{VC Synthesised Audio}. The spectrogram reveals that VC Synthesised Audio more accurately matches the original, with a stable waveform, in contrast to the stretched, high-energy T5 Synthesised Audio, which deviates significantly from the reference.

\subsection{MOS and Recognition Accuracy Calculation Protocol}\label{a.4}
We conducted a survey where 35 candidates heard 10 sets of voice clip. One set included the ground truth (original audio) and the corresponding generated audio. 

\textbf{MOS Score:} Once they heard a pair, the candidates filled out a questionnaire with 5 questions on a 1-5 rating scale. 
\begin{itemize}
    \item Rate the naturalness of the clip (assessment of non-robotic voice)
    \item Rate the emphasis and intonation of spoken words
    \item Does the speaker of the two clips sound the same
    \item Evaluate rhythm and speech consistency
    \item Fluency of speech of the generated clip versus the input clip
\end{itemize} 

\textbf{Recognition Accuracy:} This evaluates if the clips can be identified as the same user. Yes or no response.

\end{document}